\documentclass[runningheads]{llncs}
\usepackage[T1]{fontenc}
\usepackage{graphicx}
\usepackage{hyperref}
\usepackage{color}

\usepackage{enumitem}
\usepackage{url}
\usepackage{graphicx}
\usepackage{amsmath}
\usepackage{booktabs}
\usepackage{amssymb}
\usepackage{adjustbox}
\usepackage{multirow}
\graphicspath{ {./img/} }
\usepackage{algorithm}
\usepackage{algpseudocode}
\algrenewcommand\textproc{}
\usepackage{array}
\newcolumntype{H}{>{\setbox0=\hbox\bgroup}c<{\egroup}@{}}

\usepackage{pifont}
\newcommand{\cmark}{\ding{51}}
\newcommand{\xmark}{\ding{55}}

\makeatletter
\newcommand\footnoteref[1]{\protected@xdef\@thefnmark{\ref{#1}}\@footnotemark}
\makeatother

\begin{document}
\title{Combining Data Generation and Active Learning for Low-Resource Question Answering}
\titlerunning{Data Generation with AL for Low-Resource Question Answering}
%
\author{Maximilian Kimmich\thanks{Work partially done during an internship at IBM Research, Zurich.}\inst{1} \and
  Andrea Bartezzaghi\inst{2} \and
  Jasmina Bogojeska\inst{3} \and
  Cristiano Malossi\inst{2} \and
  Ngoc Thang Vu\inst{1}}
\authorrunning{M. Kimmich et al.}
%
\institute{
  University of Stuttgart, Stuttgart, Germany\\
  \email{\{maximilian.kimmich,thang.vu\}@ims.uni-stuttgart.de} \and
  IBM Research Zurich, Zurich, Switzerland\\
  \email{\{abt,acm\}@zurich.ibm.com} \and
  Centre for AI, Zurich University of Applied Sciences, Winterthur, Switzerland\\
  \email{jasmina.bogojeska@zhaw.ch}
}
\maketitle              
%
\begin{abstract}
  Neural approaches have become very popular in Question Answering (QA), however, they require a large amount of annotated data.
  In this work, we propose a novel approach that combines data augmentation via question-answer generation with Active Learning to improve performance in low-resource settings, where the target domains are diverse in terms of difficulty and similarity to the source domain.
  We also investigate Active Learning for question answering in different stages, overall reducing the annotation effort of humans.
  For this purpose, we consider target domains in realistic settings, with an extremely low amount of annotated samples but with many unlabeled documents, which we assume can be obtained with little effort.
  Additionally, we assume a sufficient amount of labeled data from the source domain being available.
  We perform extensive experiments to find the best setup for incorporating domain experts.
  Our findings show that our novel approach, where humans are incorporated in a data generation approach, boosts performance in the low-resource, domain-specific setting, allowing for low-labeling-effort question answering systems in new, specialized domains.
  They further demonstrate how human annotation affects the performance of QA depending on the stage it is performed.
\end{abstract}
\section{Introduction}
Machine Reading Question Answering (MRQA) is a challenging and important problem.
Facilitating targeted information extraction from documents, it allows users to get fast, easy access to a vast amount of documents available.
MRQA models generally need plenty of annotations \cite{rajpurkar_squad_2016}, therefore several methods have been devised for augmenting data by generating new annotated samples, with the ultimate goal of improving quality of predictions.
Some of these approaches show a real benefit in the downstream MRQA task; however, there is no work employing Language Models (LM) fine-tuned for generating question-answer pairs in low-resource, domain-specific settings, which are often observed in practice.

In these cases, most of the times only few annotated samples are available due to the lack of resources like domain experts \cite{otegi_conversational_2020}.
Also, since the specialized domain being usually vastly different from the publicly available labeled data of the source domain, this data is only useful to a limited extent.
Moreover, labeling is expensive, as it requires a significant amount of time from domain experts.

One approach to reduce the annotation effort and to employ humans more efficiently is to utilize Active Learning (AL) \cite{settles_active_2012}.
For this, samples are carefully selected for annotation to reduce the amount of redundant information.

Another common method for reducing the annotation effort is data generation.
By generating synthetic data, several works have shown to successfully counteract the low-resource setting \cite{alberti_synthetic_2019,shakeri_end--end_2020,shakeri_towards_2021}.
Consequently, combining data generation with AL can be another step towards reducing the high annotation effort for MRQA.
However, so far it stays unclear how few samples from the target domain, selected based on various criteria, affect the performance of the MRQA model when they are used in the training of the data generation model as well.
In this work, we introduce a novel approach that follows this idea, and investigate how employing AL, either at the data generation model or at the MRQA model, affects the overall performance of the MRQA task.
To this end, we consider collecting and annotating a small set of samples (e.g., 200), accurately selected to boost performance of the MRQA model, feasible and not too expensive.

\begin{figure}[tb]
  \centering
  \includegraphics[width=.8\columnwidth,keepaspectratio,trim=0 0 0 0,clip]{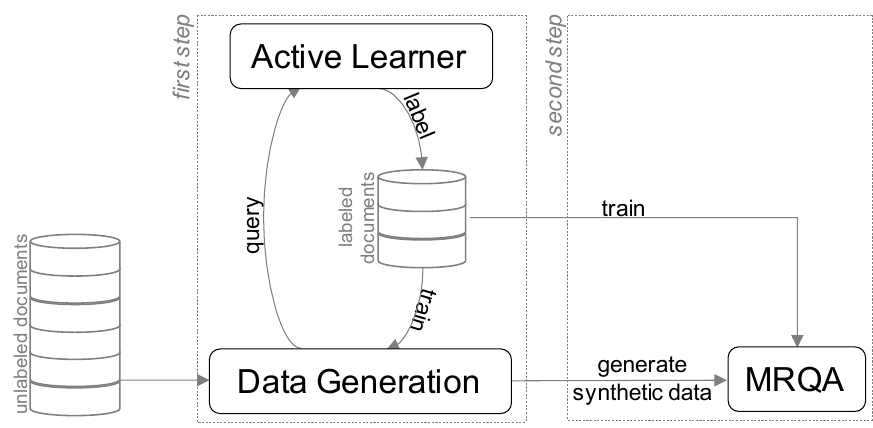}
  \caption{
    Our approach combining Active Learning with data generation: In a first step, the data generation model is efficiently trained using Active Learning. Second, this model is then used to generate data for MRQA.
  }
  \label{fig:method_overview}
\end{figure}

More precisely, we aim to answer the following research questions:
\begin{enumerate}[align=parleft, wide]
  \item[\textbf{RQ1:}] How can we combine data generation with Active Learning to improve the performance of extractive MRQA?
  \item[\textbf{RQ2:}] How can we select samples for annotation for extractive MRQA when data generation is used?
\end{enumerate}

To achieve this, we consider only a small set of documents from the \textit{target} domain annotated in terms of questions and answers, in addition to a potentially large set of samples from a different, generic \textit{source} domain.
We further assume to have access to a large amount of unlabeled data (documents) available from the target domain, which is common in real scenarios (e.g., publications from PubMed\footnote{\url{https://pubmed.ncbi.nlm.nih.gov}} when considering the medical domain).
We then employ a state-of-the-art model for generating question-answer pairs on these documents from the target domain, to enrich our training sets with synthetic data.
In our research, we study the application of AL in order to label those samples which are most relevant for increasing performance, with the aim of reducing the amount of labeled samples needed.
We consider these labeled samples in both the data generation stage as well as for the MRQA model itself.

Our contributions in this paper can be summarized as follows:
1) We introduce a novel approach that combines data augmentation for MRQA via question-answer generation with Active Learning; 2) we identify the most relevant samples for AL by adapting to our setting scoring functions recently used for unsupervised quality assessment of machine translation; 3) we introduce a new sample relevance score specific to MRQA by coupling the generated samples with the downstream task, so that the MRQA model influences the data augmentation process as well; and 4) we perform extensive experiments\footnote{Code is available at \url{https://github.com/mxschmdt/mrqa-gen-al}} to demonstrate that AL is best used in our setup at the stage of data generation for improving MRQA performance in low-resource, domain-specific scenarios.

Our experimental results show great improvements for the extractive MRQA task considering two domain-specific datasets, namely TechQA \cite{castelli_techqa_2019} and BioASQ \cite{tsatsaronis_overview_2015}, with few samples specifically annotated for the data generation model.
In numbers, combining AL with data generation increases F1 by up to 5.13\% compared to applying AL at the MRQA model.

\section{Related Work}
\subsection{Low-Resource MRQA}
In the literature, there are several approaches for tackling the low-resource problem, including scenarios where few labeled data are available and others where no labels are available at all.

One approach is to use pre-trained Language Models \cite{alberti_synthetic_2019,radford_language_2019,lewis_bart_2019} backed by Transformers \cite{vaswani_attention_2017}, which can be especially useful in cases where little or no labeled data exists and it is costly to generate more.
In the best case, the LM can be used without further fine-tuning (when the target task and domain is similar to those used in the pre-training objective).
Otherwise, if unlabeled data is available, it may be used for adaptation of the LM in a self-supervised fashion.

If the low-resource domain is accompanied by some annotated samples, weak supervision -- where external sources of information are used -- becomes relevant \cite{hedderich_survey_2021,wang_generalizing_2019}.
Moreover, data augmentation \cite{zhang_multi-stage_2020,van_cheap_2021} and LM domain adaptation \cite{nishida_unsupervised_2020,zhang_multi-stage_2020} have been shown to improve performance for MRQA.

A different approach is to use domain transfer, where a model is trained on a source domain and then adapted to a different target domain, e.g., by employing adversarial training \cite{lee_domain-agnostic_2019}.

\subsection{Data Generation}
As generative models are becoming increasingly popular, recent work has shifted from domain transfer towards data generation \cite{shakeri_end--end_2020,alberti_synthetic_2019,puri_training_2020,luo_cooperative_2021,lee_generating_2020}.
Given a passage of text, the generator model learns to output question-answer pairs.
The generation model is trained on a large amount of data from the source domain and is applied to any document in any target domain, commonly employing pre-trained LMs as decoder.

Common generation methods focus only on generating the questions and assess their performance exclusively on the generated questions using automatic metrics \cite{ushio_generative_2023,liu_asking_2020,tuan_capturing_2019,yin_summary-oriented_2021,chen_reinforcement_2020,sun_answer-focused_2018}.
Only a few approaches are based on generating full question-answer pairs, where the generated data is evaluated by means of the MRQA model.
While \cite{klein_learning_2019} and \cite{luo_cooperative_2021} only perform in-domain experiments -- where the training data and data used for generating new samples come from the same domain -- \cite{shakeri_end--end_2020} and \cite{lee_generating_2020} show the generalizability of their data generation models by also performing out of domain experiments.
Recent models even perform well in the few- and zero-shot setting~\cite{schmidt_prompting-based_2024}.
Since data generation can also be improved using annotated data from the target domain~\cite{schmidt_prompting-based_2024}, it is of interest to reduce the labeling effort.

\subsection{Active Learning}
Active Learning \cite{settles_active_2012,cohn_active_1996} is a framework aimed to reduce the amount of annotated data required for training machine learning models, based on iteratively selecting a set of specific samples to be labeled by a human with expert knowledge of the domain.
AL has been widely used in Deep Learning \cite{liu_survey_2022} and NLP \cite{siddhant_deep_2018,fang_learning_2017,lowell_practical_2019,chang_using_2020,lin_active_2017}, and it has been shown to be helpful for MRQA as well:
While \cite{kratzwald_learning_2020} learn whether samples are annotated automatically or manually, others make use of pool based sampling strategies, based on heuristics, to score samples for annotation \cite{lin_active_2017}.

\section{Method}

Our novel approach to tackle MRQA in low-resource settings with optimized human effort is based on combining AL with data generation.
In this section we describe our method including the models involved, i.e., the data generation and the MRQA model (section \ref{ssec:method_data_generation}), as well as our iterative sample selection method (section \ref{ssec:method_active_learning}).
The latter is implemented using Active Learning, which adapts and combines various agnostic scores, with the addition of a novel MRQA-tailored one, to improve the annotation efficiency.
A high-level overview of our setup is provided in figure~\ref{fig:method_overview}.

\subsection{MRQA with Data Generation}
\label{ssec:method_data_generation}


Our methodology refers to the question as $q$, the answer as $a$ and the corresponding context as $c$.
Regarding data generation, we consider the \textit{QA2S} model proposed by \cite{shakeri_end--end_2020}, since this model shows best overall performance in their and our experiments.
This model makes use of a pre-trained encoder-decoder LM and is fine-tuned to generate the question given the context, followed by generating the answer given the context and the previously decoded question in a subsequent decoding step.
More details can be found in the original paper \cite{shakeri_end--end_2020}.

For the MRQA model, we employ pre-trained BERT \cite{devlin_bert_2019} for encoding the question concatenated with the context.
On top, a span extraction head models the probability for each context token to be the start and the end of the answer span.
We refer to \cite{devlin_bert_2019} for more details.

\subsection{Active Learning}
\label{ssec:method_active_learning}

\begin{algorithm}[bt]
  \caption{Model Training with AL}
  \label{alg:al_training}
  \begin{algorithmic}[1]
    \Require
    \Statex unlabeled target domain documents $\mathcal{D}_\textrm{pool}$
    \Statex labeled target domain data $\mathcal{D}_\textrm{dev}$
    \Statex scoring function $f_\textrm{score}$ taking $(\textrm{sample}, \textrm{models})$ as input
    \Statex iterations $r$, number of samples $n$
    \State Initialize involved models $\mathcal{M}$ with pre-trained weights
    \State $\mathcal{D}_\textrm{train} = \emptyset$
    \For {$\textrm{iteration}=1,2,\ldots,r$}
    \State $\mathcal{D}_\textrm{selected} \gets$ select $n$ top scoring samples from $\mathcal{D}_\textrm{pool}$ using $f_\textrm{score}(\cdot, \mathcal{M})$
    \State $\mathcal{D}_\textrm{labeled} \gets$ \textrm{label\textsuperscript{1}} $\mathcal{D}_\textrm{selected}$
    \State $\mathcal{D}_\textrm{train} \gets \mathcal{D}_\textrm{train} \cup \mathcal{D}_\textrm{labeled}$
    \State $\mathcal{D}_\textrm{pool} \gets \mathcal{D}_\textrm{pool} \setminus \mathcal{D}_\textrm{selected}$
    \State Train $\mathcal{M}$ using $\mathcal{D}_\textrm{train}$ and $\mathcal{D}_\textrm{dev}$
    \EndFor
  \end{algorithmic}
  {
  \footnotesize
  \textsuperscript{1}In our experiments we employ an oracle, i.e., we take the label from the dataset
  }
\end{algorithm}

We consider pool based sampling for our AL scenario, where samples are iteratively selected for annotation.
Therefore, a pool of unlabeled samples is ranked using a scoring function based on context $c$.
Our novel approach, combining data generation together with AL, is briefly described in Algorithm~\ref{alg:al_training}.
In all cases, we initially fine-tune the models on the generic source domain to have a grounded starting point for AL.

We borrow Sentence Probability (SP), Dropout-based Sentence Probability (D-SP) and Dropout-based Lexical Similarity (LS) from unsupervised quality estimates \cite{fomicheva_unsupervised_2020,xiao_wat_2020} and adapt them as scoring functions for sample selection using the data generation model in our setting.
While the first two functions measure the model's entropy, the idea of the LS scoring function is that the difference of likely answers can also expose the model's uncertainty.
We also introduce round-trip scoring (RT), which incorporates the MRQA model in the sample selection process:
We believe that, during data generation, it is important to link the sample selection process with the eventual downstream task in order to generate high quality samples.
Otherwise, the data generation model is not constrained in terms of questions being generated, potentially ending up generating questions which do not take the given context or answer into account or which do not add benefit to the MRQA model in general.

\subsubsection{Scoring functions}
\paragraph{Sentence Probability (SP)}
This scoring function is based on the probability distribution of the data generation model.
The contexts are scored according to the sentence probability of the answer being generated:
\begin{equation}
  \textrm{SP}(\theta) = \frac{1}{T}\sum\limits_{t=1}^T \log p(a_t|a_{<t},c,q,\theta) .
\end{equation}
We generate the question and answer using beam search of size 10 without sampling, but only use the generated answer for scoring a sample's context.

\paragraph{Dropout-based Sentence Probability (D-SP)}
In contrast to SP, D-SP makes use of multiple data generation models to compute the sentence probability:
\begin{equation}
  \textrm{D-SP} = \frac{1}{N}\sum\limits_{n=1}^N \textrm{SP}(\theta_n) .
\end{equation}
We employ dropout at inference time (in addition to during fine-tuning) to realize different subsets of the model, as described by \cite{gal_dropout_2016}, running $N$ forward passes.
The question and answer is decoded once similarly to SP and used in all forward passes.

\paragraph{Dropout-based Lexical Similarity (LS)}
Regarding LS, multiple answers are decoded using multiple models (again realized via dropout), and all of them are compared pairwise at the lexical level using the Meteor \cite{banerjee_meteor_2005} metric (with $i \neq j$):
\begin{equation}
  \textrm{LS} = \frac{1}{N*(N-1)}\sum\limits_{i=1}^N\sum\limits_{j=1}^N \textrm{Meteor}(a_i, a_j) .
\end{equation}
Decoding is again implemented using beam search of size 10 without sampling.

\paragraph{Round-trip (RT)}
Since the ultimate goal is to improve the performance of the MRQA model, we propose this novel method integrating the MRQA model in the computation of the ranking score:
Question-answer pairs are generated from the context similarly to SP, and then we apply an MRQA model in order to rank the samples according to the F1 score between the predicted answer and the answer generated by the data generation model.
We therefore prefer documents for which the data generation model will generate question-answer pairs which the MRQA model cannot predict correctly.

\section{Experimental Setup}
In this work, we focus on the low-resource setting, where we assume to be able to use an extremely small amount of 200 annotated samples from the target domain.
In addition to a large annotated source domain, we assume to have many unlabeled documents available from the target domain as well, which can be used for generating labeled samples.
We consider this assumption to be valid in general, as one of the main goals of MRQA is to provide better access to the vast amount of existing documents.

\subsection{Data}
We consider SQuAD \cite{rajpurkar_squad_2016} as source domain and NaturalQuestions (NQ) \cite{kwiatkowski_natural_2019}, TechQA \cite{castelli_techqa_2019} and BioASQ \cite{tsatsaronis_overview_2015} as target domains; statistics on tokens are summarized in table~\ref{tab:domain_stats}.
The main reason for choosing these datasets is to consider examples of very specialized domains (TechQA, BioASQ) completely different from the source domain, as well as an example of a domain overlapping with the source domain (NQ).

In our experiments  with AL, we use SQuAD for initially fine-tuning the data generation model.
Since BioASQ only includes a dev set, we create random splits for training, development and testing with 70\%, 20\% and 10\% of samples, respectively.

Regarding the documents needed for generating question-answer pairs, we use the documents from the training data in case of SQuAD and NQ.
For TechQA and BioASQ, we use passages from the corpus of technotes containing 800K+ documents and crawled abstracts from PubMed\footnote{\url{https://pubmed.ncbi.nlm.nih.gov}}, respectively.

\begin{table}[tb]
  \setlength{\tabcolsep}{8pt}
  \caption{Statistics about the train split for each domain obtained with the Bert tokenizer.}
  \label{tab:domain_stats}
  \centering
  \adjustbox{max width=\columnwidth}{%
    \begin{tabular}{l|l|l|l|l|l|l|l|l|l}
      \toprule
      \multirow{2}{*}{Domain} & \multicolumn{3}{c|}{Context tokens} & \multicolumn{3}{c|}{Questions tokens} & \multicolumn{3}{c}{Answer tokens}                                         \\
      \cmidrule{2-10}
                              & Min                                 & Max                                   & Mean                              & Min & Max & Mean  & Min & Max & Mean  \\ \midrule
      SQuAD                   & 25                                  & 853                                   & 155.75                            & 1   & 61  & 12.29 & 1   & 68  & 4.23  \\
      NQ                      & 10                                  & 3143                                  & 245.4                             & 7   & 29  & 9.76  & 1   & 270 & 5.25  \\
      TechQA                  & 38                                  & 38925                                 & 1484.08                           & 6   & 561 & 69.37 & 5   & 545 & 98.76 \\
      BioASQ                  & 27                                  & 960                                   & 337.71                            & 5   & 36  & 15.06 & 1   & 76  & 4.42  \\ \bottomrule
    \end{tabular}
  }
\end{table}

\subsection{Data Generation}
\subsubsection{Training with Active Learning}
\label{ssec:experimental_setup_active_learning}
For training the data generation model with AL, we score available unlabeled documents for annotation using each of the scoring functions described in Section~\ref{ssec:method_active_learning}, running 10 forward passes for D-SP and LS.
We only score documents for which labels (i.e., questions and answers) exist as we simulate the domain expert.
That is, for the selected unlabeled documents after scoring, we take the labels from the provided datasets (i.e., we use them as an oracle).
We run 4 iterations, starting initially with all available samples from the target domain in the pool, and then selecting the 50 samples the model is least confident about in each step.
In each iteration, we remove the selected samples from the query pool and fine-tune the model on all samples selected so far, always re-initializing the model with weights gained when fine-tuning on the source domain in order to not introduce any bias by having the model see some samples more often.
Since the MRQA model is used in RT, we also train the MRQA model in each iteration when this scoring functions is used.

Furthermore, as D-SP, LS and RT are computationally expensive, with LS actually decoding generated text after each forward pass, performing AL with LS and RT on NQ was not feasible for us.
Therefore, we selected 10000 random samples from this domain for use with AL, denoted as NQ\#10000 in our experiments.

Regarding the models, we use \texttt{bart-large} \cite{lewis_bart_2019} with 406M parameters for the data generation model as this showed better results in our preliminary experiments compared to \texttt{gpt2-large} \cite{radford_language_2019} and is similar to previous work~\cite{shakeri_end--end_2020} from where we also take hyperparameters.
Therefore, the data generation model is fine-tuned using cross entropy loss with AdamW Optimizer \cite{loshchilov_decoupled_2019}, a learning rate of $3\mathrm{e}{-5}$ with $10\%$ warm-up and a batch size of $24$.
The best model is selected on the evaluation data w.r.t. loss.
Fine-tuning of the data generation model is performed for 5 epochs on source domain data first, followed by fine-tuning on the target domain.
As we found that 5 epochs is too few for fine-tuning in the low-resource setting, on the target domain we fine-tune it for 10 epochs instead.
We use this type of large language model as it can be trained with rather low hardware constraints, and iterative training is crucial to AL.
For details of the MRQA model and its training see section \ref{ssec:experimental_setup_mrqa_model}.

To fit the context concatenated with the question into the models, we split the context into several chunks.
We only consider those chunks where the answer occurs in the context.
Additionally, since TechQA has long questions, we truncate the questions to the first 200 tokens, to allow for sufficient space in the input for the context as well.

\subsubsection{Synthetic Data Generation}
Regarding question generation (i.e., the first decoding step), we allow contexts to have a size of up to 724 tokens; this allows the generated questions to be composed of up to 300 tokens, as both will be fed into the model in the second decoding step and their sum cannot exceed the total of 1024 tokens (including special tokens).
Chunking this input would be difficult, since aggregating the generated answer is not simple (especially if outputs do not overlap).
We generate 10 questions per context, followed by one answer for each question-context pair, yielding up to 10 samples per context.
We do this to increase the variance of generated data for which different questions are crucial, but there can only be a single answer for each question-context pair.

Question generation is performed using nucleus sampling \cite{holtzman_curious_2020}, considering 95\% of the probability mass (top-p) and top 20 tokens (top-k) in each step.
The answer is then decoded using beam search of size 10 without sampling similar to prior work~\cite{shakeri_end--end_2020}.
Samples for which the answer does not occur in the context are discarded since we focus on extractive MRQA.
Furthermore, only samples for which the corresponding end tokens are predicted correctly are considered as valid.

In total, we choose 100000 documents from the target domain, ignoring contexts with less than 100 tokens.
In case of NQ, only 50535 documents remain.

\subsubsection{Filtering}
Finally, only a subset of the generated samples is kept.
Therefore, they are filtered using two sample filtering approaches to get rid of noisy samples, namely the LM score filtering \cite{shakeri_end--end_2020} and the round-trip consistency (RTcons) \cite{alberti_synthetic_2019}.
In LM score filtering, the generated samples are sorted according to the probability $p(y|y_{<t},x)$ as given by the generation model and the top $n$ samples are kept (we use $n=5$).
Regarding the RTcons filtering, an MRQA model is used to assess the generated question-answer pair:
The generated question along with the context is fed into the MRQA model to predict the answer.
The sample is then discarded if the predicted answer does not match the generated one.
Since RTcons might discard all samples generated from a context, we fine-tune the MRQA model (previously fine-tuned on the source domain) used for RTcons filtering also on the available samples from the target domain.

\subsection{MRQA Model}
\label{ssec:experimental_setup_mrqa_model}
In all cases (i.e., for scoring with RT, for filtering with RTcons and for the eventual downstream task), we use \texttt{bert-base-uncased} \cite{devlin_bert_2019} for the MRQA model similar to the setup in \cite{shakeri_end--end_2020}.
For both, training and inference, the following configuration applies:
A maximum input length of 512 tokens and a context stride of 128 is used.
Also, similar to the data generation model, when considering the TechQA dataset, we truncate the questions in the input.

Similar to the data generation model, we borrow hyperparameters from previous work~\cite{shakeri_end--end_2020}, therefore training the MRQA model with AdamW Optimizer \cite{loshchilov_decoupled_2019}, a learning rate of $3\mathrm{e}{-5}$ and a batch size of $24$.
Warm-up was disabled for fine-tuning the MRQA model.

In case of fine-tuning on target domain data, model training turned out to be quite fluctuating in terms of the evaluation score.
Therefore, we performed a hyperparameter search w.r.t. F1 score including learning rate, warm-up steps (using linear scheduler), L2 regularization, pre-trained weight decay for the encoder and for the output layer separately, freezing the encoder or its embedding, training of only top-n layers and re-initializing top-n layers.
As a result of manual tuning, only pre-trained weight decay on the encoder with $\lambda=1\mathrm{e}{-7}$ was employed, while all layers were fine-tuned without re-initialization.

Since chunking is applied to the inputs and predictions are aggregated afterwards (by choosing the best span over all chunks' predictions), it may happen that the answer of a sample is not completely included in a chunk's context.
We especially observe this for TechQA where answers tend to be longer, as the performance on the evaluation data is greatly degraded as the model never has the chance to predict the answer correctly for some samples.

For the downstream task, we have seen better results in preliminary experiments when first fine-tuning on generated data followed by fine-tuning on labeled target domain data in a subsequent step.

\subsection{Baselines}
In order to assess the performance of our approach, we compare it against several baselines.
Since we aim at comparing the use of humans at different stages within the process of model development instead of solely improving MRQA performance, we compare our approach to other, effective approaches utilizing humans ($b_3$, $b_5$) while baselines $b_1$, $b_2$ and $b_4$ serve as a measure of human impact.

In absence of AL, our first baseline ($b_1$) consists of an MRQA model trained only on randomly selected labeled target domain data (again 200 samples).
In addition, $b_2$ is additionally trained on SQuAD first.

The last baseline without data generation ($b_3$) differs from $b_2$ in that it does not use random target domain data but rather employs AL.
For this purpose, we consider Bayesian Active Learning by Disagreement (BALD) \cite{houlsby_bayesian_2011,gal_deep_2017} to score target domain samples.
BALD measures the uncertainty of the model's predictions, using entropy $\textrm{H}$ given training data $\mathcal{D}$ and model weights $\theta$:
\begin{equation}
  \textrm{H}[y|x,\mathcal{D}] - \mathbb{E}_{\theta\sim p(\theta|\mathcal{D})}\textrm{H}[y|x,\theta] .
\end{equation}
In order to make computation feasible, we calculate the BALD score (using dropout) independently for start and end probabilities.
The sum of both scores is then used to rate and order the unlabeled samples where higher scores are preferred.

To research our approach employing AL at the data generation model, we also consider two baselines where data generation is used.
While $b_4$ selects samples randomly, $b_5$ again selects samples using BALD after synthetic data generation.

For reference, we also report results of MRQA models only trained on synthetic data involving SQuAD training data only ($r_1$) as well as on the full set of target domain training samples ($r_2$).

In all cases (but $r_2$), if target domain data is used, a total of 200 labeled target domain samples is selected.

\section{Results \& Discussion}

\begin{table*}[tb]
  \caption{%
    F1 (std.~dev. over 5 model runs) for different AL approaches for MRQA with and without data generation showing only best filtering strategy (for synthetic data) for each model ($^\ast$ for LM filtering, $^\dagger$ for RTcons filtering).
    Without data generation only the MRQA model is used.
    AL improves results for all domains; best overall results are highlighted in bold.
  }
  \label{tab:results}
  \centering
  \adjustbox{max width=\textwidth}{%
    \begin{tabular}{l|@{\hskip 4pt}ll@{\hskip 4pt}|H@{\hskip 4pt}cHcHcHc}
      \toprule
      \multirow{2}{*}{{\textbf{Model}}}                       &
      \multicolumn{2}{c|}{\textbf{Target}}                    &
      \multicolumn{2}{c}{\multirow{2}{*}{\textbf{NQ}}}        &
      \multicolumn{2}{c}{\multirow{2}{*}{\textbf{NQ\#10000}}} &
      \multicolumn{2}{c}{\multirow{2}{*}{\textbf{TechQA}}}    &
      \multicolumn{2}{c}{\multirow{2}{*}{\textbf{BioASQ}}}                                                                                                                                                                                                                                                                                                                                                                                   \\
                                                              & \multicolumn{2}{c|}{\textbf{Domain}\textsuperscript{1}} &                                                                                                                                                                                                                                                                                                                    \\
      \midrule
      No Data Generation                                      &                                                         &          &                                     &                                      &                                   &                                   &                                     &                                      &                                   &                                   \\
      \quad w/o SQuAD ($b_1$)                                 & \cmark                                                  & (random) & 20.61 $\pm$ 1.01                    & 29.28 $\pm$ 0.86                     & 20.61 $\pm$ 1.01                  & 29.28 $\pm$ 0.86                  & 24.75 $\pm$ 1.02                    & 53.81 $\pm$ 0.68                     & 39.93 $\pm$ 1.46                  & 46.52 $\pm$ 1.67                  \\
      \quad w/ SQuAD ($b_2$)                                  & \cmark                                                  & (random) & 49.26 $\pm$ 0.23                    & 62.72 $\pm$ 0.3                      & 49.26 $\pm$ 0.23                  & 62.72 $\pm$ 0.3                   & 27.75 $\pm$ 0.64                    & 56.19 $\pm$ 0.89                     & 67.31 $\pm$ 0.72                  & 73.96 $\pm$ 0.32                  \\
      \quad w/ SQuAD ($b_3$)                                  & \cmark                                                  & (BALD)   & 48.61 $\pm$ 0.84                    & 63.28 $\pm$ 0.68                     & 49.78 $\pm$ 0.45                  & 64.12 $\pm$ 0.42                  & 27 $\pm$ 0.47                       & 53.89 $\pm$ 0.96                     & 65.58 $\pm$ 1.16                  & 74.69 $\pm$ 1.23                  \\
      \midrule
      Data Generation                                         &                                                         &          &                                     &                                      &                                   &                                   &                                     &                                      &                                   &                                   \\
      \quad w/ SQuAD ($r_1$)                                  & \xmark                                                  &          & 51.37 $^\dagger$                    & 64.38 $^\dagger$                     & 51.37 $^\dagger$                  & 64.38 $^\dagger$                  & 0.63 $^\ast$                        & 19.76 $^\ast$                        & 50.5 $^\ast$                      & 62.99 $^\ast$                     \\
      \quad w/ SQuAD ($b_4$)                                  & \cmark                                                  & (random) & 54.61 $\pm$ 0.15 $^\ast$            & 67.13 $\pm$ 0.15 $^\ast$             & 54.61 $\pm$ 0.15 $^\ast$          & 67.13 $\pm$ 0.15 $^\ast$          & 30.63 $\pm$ 0.83 $^\ast$            & 57.04 $\pm$ 1.24 $^\ast$             & 72.36 $\pm$ 0.95 $^\ast$          & 80.15 $\pm$ 0.85 $^\ast$          \\
      \quad w/ SQuAD ($b_5$)                                  & \cmark                                                  & (BALD)   & 52.53 $\pm$ 0.44 $^\dagger$         & 66.18 $\pm$ 0.36 $^\dagger$          & 53.10 $\pm$ 0.47 $^\ast$          & 66.58 $\pm$ 0.42 $^\ast$          & 27.88 $\pm$ 1.29 $^\dagger$         & 57.04 $\pm$ 0.93 $^\dagger$          & 68.7 $\pm$ 1.02 $^\ast$           & 77.44 $\pm$ 0.53 $^\ast$          \\
      \midrule
      Data Generation using AL                                &                                                         &          &                                     &                                      &                                   &                                   &                                     &                                      &                                   &                                   \\
      \quad SP                                                & \cmark                                                  & (SP)     & 54.48 $\pm$ 0.02 $^\dagger$         & 67.23 $\pm$ 0 $^\dagger$             & 55.39 $\pm$ 0.07 $^\dagger$       & 68.04 $\pm$ 0.07 $^\dagger$       & 30.38 $\pm$ 0.85 $^\ast$            & 56.51 $\pm$ 0.73 $^\ast$             & 72.03 $\pm$ 0.77 $^\ast$          & 80.63 $\pm$ 0.59 $^\ast$          \\
      \quad D-SP                                              & \cmark                                                  & (D-SP)   & \textbf{54.8 $\pm$ 0.16} $^\dagger$ & \textbf{67.68 $\pm$ 0.11} $^\dagger$ & 55.11 $\pm$ 0.08 $^\ast$          & 67.86 $\pm$ 0.08 $^\ast$          & \textbf{32.5 $\pm$ 0.56} $^\dagger$ & 56.17 $\pm$ 0.41 $^\dagger$          & \textbf{75.35 $\pm$ 0.64} $^\ast$ & \textbf{82.57 $\pm$ 0.52} $^\ast$ \\
      \quad LS                                                & \cmark                                                  & (LS)     & -                                   & -                                    & 54.83 $\pm$ 0.19 $^\ast$          & 67.75 $\pm$ 0.16 $^\ast$          & 31.25 $\pm$ 1.25 $^\dagger$         & 55.5 $\pm$ 1.39 $^\dagger$           & 71.89 $\pm$ 0.72 $^\dagger$       & 79.95 $\pm$ 0.53 $^\dagger$       \\
      \quad RT                                                & \cmark                                                  & (RT)     & -                                   & -                                    & \textbf{55.43 $\pm$ 0.05} $^\ast$ & \textbf{68.28 $\pm$ 0.08} $^\ast$ & 32 $\pm$ 0.47 $^\dagger$            & \textbf{58.89 $\pm$ 0.62} $^\dagger$ & 71.96 $\pm$ 0.16 $^\ast$          & 80.23 $\pm$ 0.44 $^\ast$          \\
      \midrule[1.5pt]
      MRQA only w/o SQuAD ($r_2$)                             & \cmark                                                  & (Full)   & 66.63                               & 78.64                                & 66.63$\textsuperscript{2}$        & 78.64$\textsuperscript{2}$        & 26.88                               & 55.9                                 & 78.07                             & 80.64                             \\
      \bottomrule
    \end{tabular}
  }
  {
    \raggedright
    $\textsuperscript{1}$200 labeled samples from the target domain drawn according to strategy (but \textit{Full} where all target domain data is used)\\
    $\textsuperscript{2}$subsampling is only applied to AL hence this was trained on the full training data\\
  }
\end{table*}

In the following, we describe the results and answer our research questions.
Commonly used F1 score is used to evaluate the proposed approach as well as the baselines.

\subsection{RQ1: Combining AL with Data Generation}

We report results of our experiments in table~\ref{tab:results}.
In our scenario, AL performs better than random sampling, both when it is applied on the MRQA model and when it is applied on the data generation model as well, as the results obtained with our various scoring functions demonstrate.

Without AL, we get the best results for our low-resource setting on all datasets when data generation is employed using both, SQuAD and target domain data ($b_4$).
In terms of F1 metric, we get 67.13\% on NQ, 57.04\% on TechQA and 80.15\% on BioASQ.
However, we observe that AL improves performance on all domains, with an absolute increase of F1 of 0.55\% (NQ), 1.15\% (NQ\#10000), 1.85\% (TechQA) and 2.42\% (BioASQ).
If compared to applying AL at the MRQA model ($b_5$), these improvements increase to 1.5\% (NQ), 1.7\% (NQ\#10000) and 5.13\% (BioASQ).
Performance on NQ is in general not improved much when including additional data from the target domain, probably because of its similarity with the source domain (SQuAD) w.r.t.~the others.

Although fine-tuning the data generation model on random target domain data ($b_4$) gives already superior results w.r.t. BALD ($b_5$) when data generation is used, applying AL at the MRQA model still improves performance on all datasets ($b_3$, $b_5$), if compared to random target domain data with ($b_4$) and without data generation ($b_2$).
However, when applying AL at the MRQA model, a large improvement for all domains can still be observed by applying synthetic data generation ($b_5$ vs. $b_3$).

Therefore, for answering our research question RQ1, we can say that applying AL improves the MRQA task when it is combined with data generation, as our findings from AL on the data generation model in contrast to AL on the MRQA model demonstrate.
We expect this improvement to be further increased when more diverse (unlabeled) samples are available for querying.
Also, TechQA and BioASQ actually outperform the full target domain setting.
We believe this is due to the additional generated data which is useful for the target domain.
In contrast, TechQA and BioASQ are both relatively small in terms of the number of training samples.

We find that target domain data is necessary when tackling more specialized domains, as in the case of TechQA and BioASQ, as this improves F1 by up to 39.13\% and 19.58\% (compared to $r_1$), respectively, with only 200 samples annotated.

Therefore, AL combined with data generation can thus be used as an approach to minimize the costly annotation effort when targeting specialized domains.

\subsection{RQ2: Selecting Samples for AL in Data Generation}
For answering our second research question, we take a closer look at the scoring functions for data generation.
In our experiments we observe that our proposed method, RT, consistently improves all domains, outperforming random target data selection.
Additionally D-SP also shows strong results, but falls behind in case of TechQA.
We assume that the small number of documents AL can query in TechQA limits its capabilities, and that more data could lead to an improvement.
In numbers, while RT yields best F1 scores for NQ\#10000 (68.28\%) and TechQA (58.89\%), BioASQ profits the most from D-SP (82.57\%) with the latter two domains actually outperforming the case where all target domain samples are annotated ($r_2$).

In addition, we do not observe any degradation of performance when subsampling NQ.
Therefore, subsampling large pools of (unlabeled) samples for better efficiency seems to be a valid choice when applying AL.

\section{Analysis}
To better understand the performance of the AL methods considered in our data generation setup, we analyze different aspects of the selected samples.

\begin{table}[tbh]
  \setlength{\tabcolsep}{8pt}
  \caption{Context and question length in tokens on TechQA for samples selected with different strategies: AL strategies on the data generation model prefer samples with long contexts.}
  \label{tab:techqa_samples_stats}
  \centering
  \adjustbox{max width=\columnwidth}{%
    \begin{tabular}{lcllll}
      \toprule
      \multirow{2}{*}{\textbf{AL strategy}} & \multirow{2}{*}{\textbf{AL applied to}} & \multicolumn{2}{c}{\textbf{$\varnothing$ context length}} & \multicolumn{2}{c}{\textbf{$\varnothing$ question length}}                 \\
      \cmidrule(lr){3-4}\cmidrule(lr){5-6}
                                            &                                         & Bert                                                      & Bart                                                       & Bert  & Bart  \\
      \midrule
      BALD                                  & MRQA                                    & 1462.45                                                   & 1521.82                                                    & 60.43 & 67.94 \\
      SP                                    & data generation                         & 2564.99                                                   & 2715.05                                                    & 71.65 & 83.34 \\
      LS                                    & data generation                         & 2519.89                                                   & 2660.34                                                    & 69.98 & 77.63 \\
      RT                                    & data generation                         & 1836.84                                                   & 1950.62                                                    & 75.98 & 81.28 \\
      D-SP                                  & data generation                         & 2531.28                                                   & 2686.99                                                    & 69.32 & 80.47 \\
      Random                                & data generation                         & 1588.09                                                   & 1668.24                                                    & 67.79 & 74.43 \\
      \bottomrule
    \end{tabular}
  }
\end{table}
\subsection{Statistics on drawn samples}

\begin{table}[tbh]
  \setlength{\tabcolsep}{8pt}
  \caption{Overlap of samples for the examined AL strategies on BioASQ: The overlap of the scoring functions is rather small.}
  \label{tab:bioasq_samples_overlap}
  \centering
  \adjustbox{max width=\columnwidth}{%
    \begin{tabular}{lcrrrrHrr}
      \toprule
      \multirow{2}{*}{\textbf{AL strategy}} & \multirow{2}{*}{\textbf{AL applied to}} & \multicolumn{7}{c}{\textbf{overlap with AL strategy}}                                             \\
      \cmidrule(lr){3-9}
                                            &                                         & BALD                                                  & SP  & LS  & RT  & D-SP+RT & D-SP & Random \\
      \midrule
      BALD                                  & MRQA                                    & 200                                                   & 39  & 44  & 42  & 54      & 49   & 42     \\
      SP                                    & data generation                         & 39                                                    & 200 & 49  & 49  & 66      & 72   & 40     \\
      LS                                    & data generation                         & 44                                                    & 49  & 200 & 37  & 48      & 46   & 41     \\
      RT                                    & data generation                         & 42                                                    & 49  & 37  & 200 & 63      & 49   & 15     \\
      D-SP                                  & data generation                         & 49                                                    & 72  & 46  & 49  & 82      & 200  & 36     \\
      Random                                & data generation                         & 42                                                    & 40  & 41  & 15  & 42      & 36   & 200    \\
      \bottomrule
    \end{tabular}
  }
\end{table}

Applying AL on the data generation model selects samples with long contexts for all domains.
Exemplary statistics are shown in table~\ref{tab:techqa_samples_stats} for TechQA.
RT selects samples with long contexts.
Still, they are in average much shorter then for the other AL scoring functions for the data generation model.
Furthermore, AL on the MRQA model using BALD does not put emphasis on long contexts which can further explain the improvements of applying AL at the stage of data generation.

\subsection{Overlap of drawn samples}
We analyze the overlap of the chosen samples between the different AL strategies.
Table~\ref{tab:bioasq_samples_overlap} reflects this for the BioASQ domain.
We observe rather small sample overlap among the approaches, a trend that holds for the other domains as well.
The overlap depends on the dataset with larger datasets having less overlap.

\subsection{Distribution of scores}

\begin{figure}[tbh]
  \centering
  \includegraphics[clip,trim=0cm 0cm 2cm 3.2cm,width=.9\linewidth,keepaspectratio]{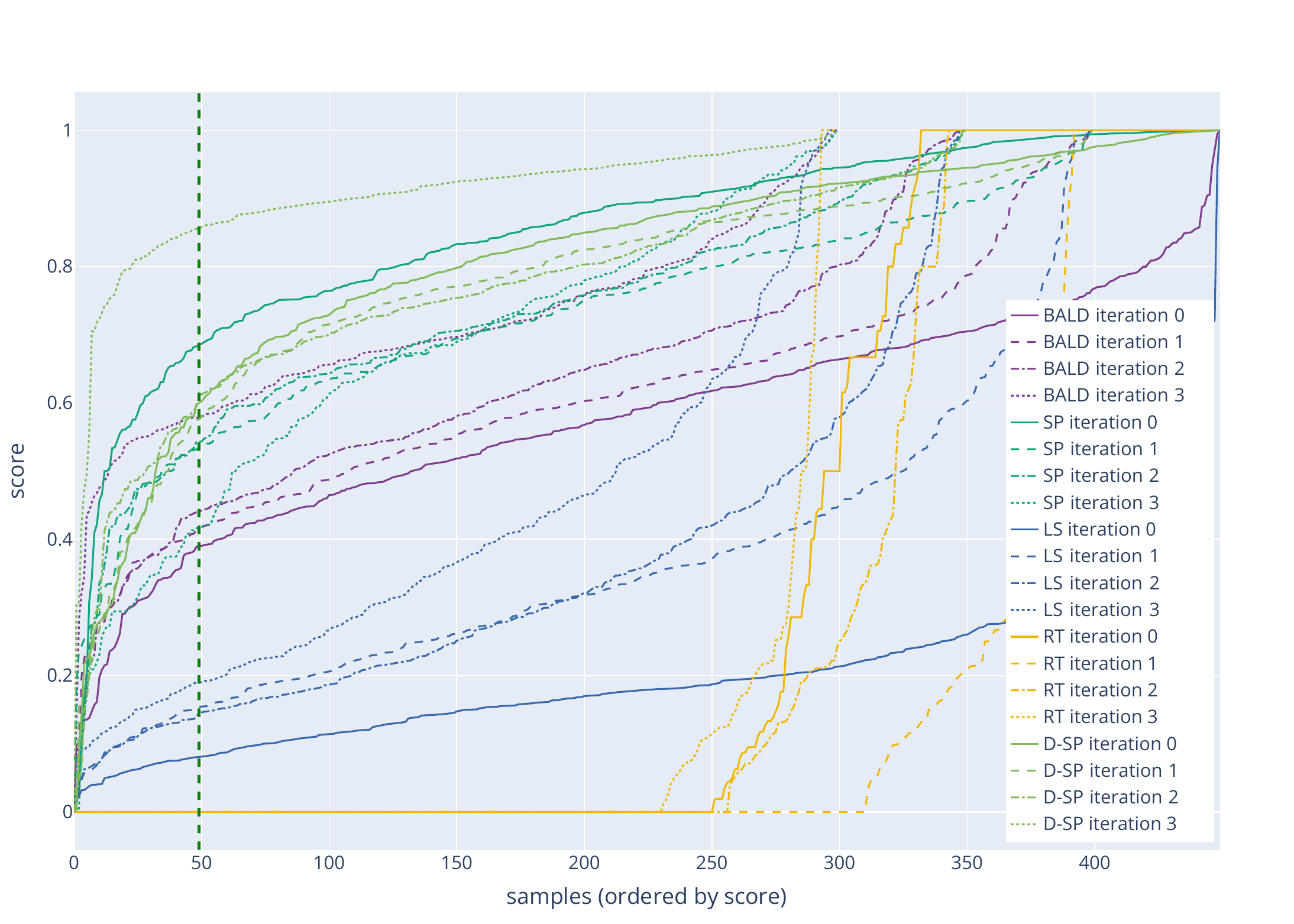}
  \caption{Sample score distribution for TechQA: RT scores many samples low, but surprisingly also rates some samples high, although the task of predicting the generated answer for a generated question is complex. Scores have been rescaled to $\left[0,1\right]$ per scoring function and iteration to better compare distributions.}
  \label{fig:sample_scores_techqa}
\end{figure}

Figure~\ref{fig:sample_scores_techqa} shows the distribution of scores over all available samples in each iteration of AL for the TechQA dataset.
Regarding RT scoring, we observe a steep curve where, depending on the dataset, many samples are rated as 0 and many as 1.
This indicates that both models, data generation and MRQA, work very well in combination since well working samples are rated high.
The RT scoring seems to be also a good measure to quantify the ``difference'' between the target domain and the source domain.
For example, for both NQ and BioASQ, RT scores are much better than those for TechQA.
This happens probably because the RT scoring method considers the downstream MRQA task when computing the score thus reflecting the larger distribution shift when more specialized domains are considered.
Therefore, for the TechQA dataset, the visualization suggests that one might as well select all 200 samples in one iteration, instead of annotating 50 samples in four consecutive iterations.
Another interesting behavior we observe is that the RT method scores samples best in the first iteration.
This might occur due to potential overfitting of the data generation model, being fine-tuned with 50 samples from the target domain, resulting in a decrease of score in the following iteration.

\subsection{Distribution of samples}

\begin{figure}[tbh]
  \centering
  \includegraphics[clip, trim=0.65cm 0.85cm 0.65cm 3.9cm, width=.8\linewidth]{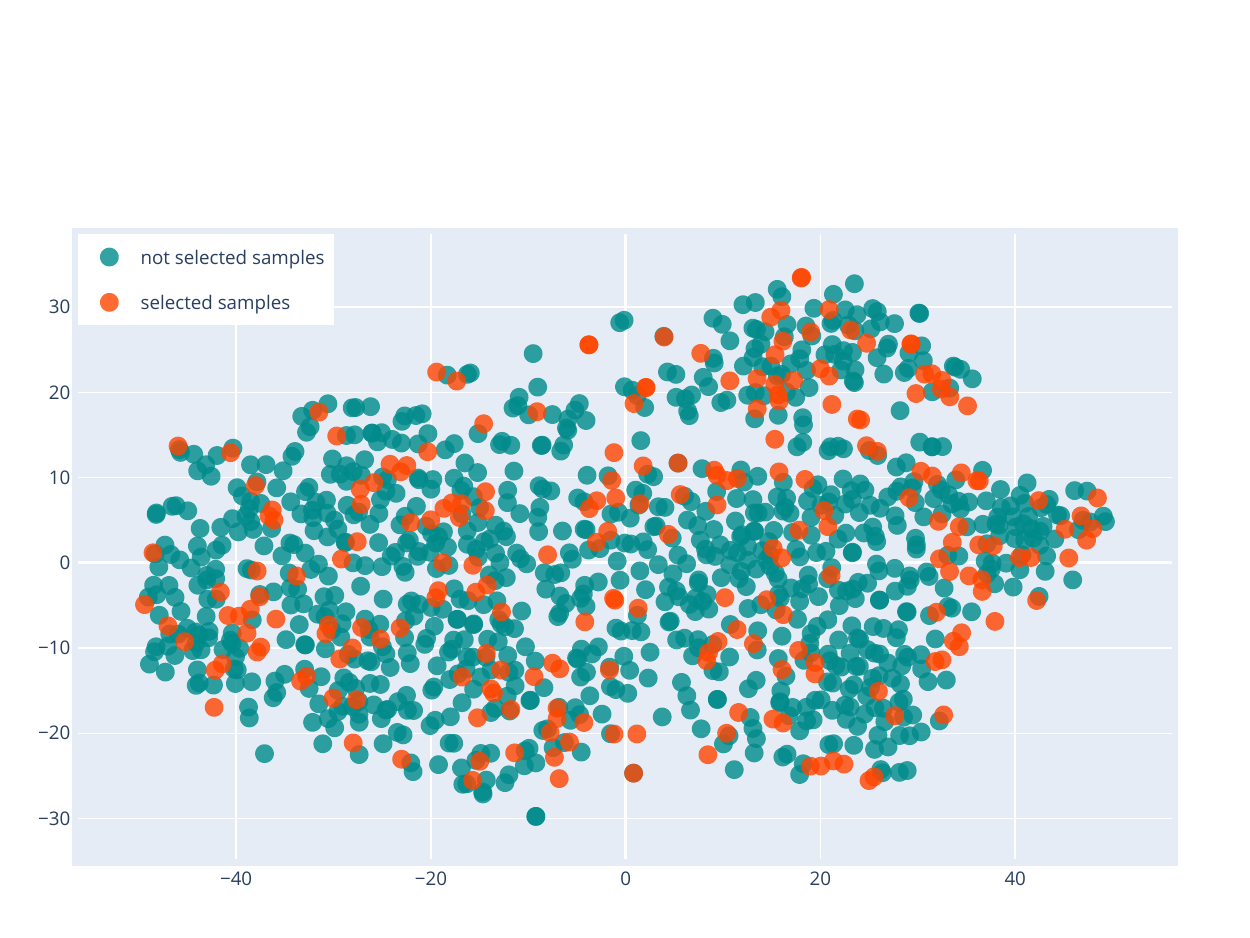}
  \caption{Visualization of the BioASQ dataset samples using representations retrieved using the MRQA model, with samples selected by RT in the last iteration (221 instances) marked red. Selected samples are well distributed among all samples suggesting that a diverse set of samples is selected.}
  \label{fig:samples_dist_bioasq_rt_mrqa_last_iteration}
\end{figure}

In order for the MRQA model to perform well, it is important that a set of diverse samples is selected by the AL strategy.
Figure~\ref{fig:samples_dist_bioasq_rt_mrqa_last_iteration} shows a visualization, based on t-SNE, of the diversity of samples drawn from BioASQ according to RT scoring in the last iteration.
We find that selected samples are well distributed among the available samples.

\section{Conclusion}
Although data scarcity is ubiquitous in practice, there is a lack of approaches that address this issue for the MRQA task in specialized domains.
Therefore, appropriate methods applicable to realistic low-resource domains are of great importance since deep learning models usually work well when plenty of annotated, in-domain samples are available.
Our work shows how to utilize Active Learning to boost performance when addressing the challenging MRQA problem in low-resource, domain-specific scenarios relevant in practice.
To this end, we introduced a novel approach that combines data generation with AL, resulting in a boost in performance with low labeling effort.
We further analyzed the performance of AL when applied to the MRQA model directly compared to when applied to the data generation process, showing evidenced improvements in predictive performance in the latter scenario, especially when using our newly introduced scoring function tailored to the MRQA task.

\begin{credits}
  \subsubsection{\ackname} This work is supported by IBM Research AI through the IBM AI Horizons Network.

\end{credits}
%
%
%
\bibliographystyle{splncs04}
\bibliography{main}
\end{document}